\pdfoutput=1

\documentclass[11pt]{article}

\usepackage[preprint]{acl}

\usepackage{times}
\usepackage{latexsym}
\usepackage{microtype}
\usepackage{graphicx}
\usepackage{subfigure}
\usepackage{booktabs} 
\usepackage{listings} 
\usepackage{graphicx}
\usepackage[table,xcdraw]{xcolor}
\usepackage{multirow}

\usepackage[T1]{fontenc}

\usepackage[utf8]{inputenc}

\usepackage{microtype}

\usepackage{inconsolata}

\usepackage{graphicx}

\usepackage{amsmath}
\usepackage{amssymb}
\usepackage{mathtools}
\usepackage{amsthm}

\usepackage[capitalize,noabbrev]{cleveref}

\theoremstyle{plain}
\newtheorem{theorem}{Theorem}[section]

\theoremstyle{definition}
\newtheorem{definition}[theorem]{Definition}

\theoremstyle{remark}

\title{Efficient Embedding-based Synthetic Data Generation for Complex Reasoning Tasks}

\author{
Srideepika Jayaraman,  Achille Fokoue,  Dhaval Patel,  Jayant Kalagnanam \\
IBM Research, Yorktown Heights, NY, USA \\
\texttt{j.srideepika@ibm.com, \{achille, pateldha, jayant\}@us.ibm.com}
}

\begin{document}
\maketitle

\begin{abstract}
Synthetic Data Generation (SDG), leveraging Large Language Models (LLMs), has recently been recognized and broadly adopted as an effective approach to improve the performance of smaller but more resource and compute efficient LLMs  through fine-tuning. A key challenge in SDG is ensuring the quality and diversity of the generated data. In this paper, we analyze the diversity and distribution of generated data in the embedding space, and demonstrate a strong correlation between the density of examples within a specific neighborhood and the accuracy of predictions on examples drawn from that region. Building on this insight, we present a targeted pipeline for embedding-based sampling that enhances data diversity and consistently improves performance across several benchmarks. 

\end{abstract}
\section{Introduction}
\label{intro}
In recent years, Large Language Models (LLMs) have dramatically improved machines’s ability to understand and generate natural language. The rapid growth in size of the most capable LLMs has raised serious concerns about their resource consumption and sustainability. As a result, there have been increased research efforts in exploring approaches to bring the performance of much smaller LLMs (less than 20B parameters) closer to the performance of larger models (100B+ parameters).  Synthetic Data Generation (SDG) has recently been recognized and broadly adopted as one effective approach to improve the performance of smaller, more resource and compute efficient LLMs through fine-tuning.

Synthetic Data Generation (SDG) is typically a model distillation approach that uses a more capable teacher model to generate synthetic training examples used to then trained or fine-tuned a smaller LLM on a specific set of tasks. A key challenge in SDG is ensuring the quality and diversity of the generated data.  Most SDG techniques generate new synthetic examples by sampling seed examples from an existing set of known training examples (hereafter referred to as the pool of seed examples).  Unfortunately, most prior works (e.g., ~\cite{taori2023stanford, wang2022self}) often rely on random sampling of the pool of seed examples, which, as explained by ~\cite{gudibande2023false,sudalairaj2024lab},  leads to over-sampling from the dominant modes of the teacher model - resulting in limited diversity. ~\cite{sudalairaj2024lab} tackles this issue by proposing a new approach that first requires manually building a taxonomy and placing all examples in the pool of seed examples in the taxonomy. It then performs stratified sampling through the taxonomy (taxonomy-driven sampling).  However, the success of the approach depends on the existence of a well-designed, well balanced and well-organized taxonomy and the proper mapping of examples in the pool of seed examples to the appropriate nodes in the taxonomy.

As opposed to prevailing SDG approaches that study, organize and sample the pool of seed examples in the language domain, we propose to study, organize and sample it in an embedding space. Furthermore, while prior SDG works have paid limited to no attention to the target (or student) model that will eventually be fine-tuned on the synthetically generated data, in this paper, we design the SDG process to specifically overcome the shortcomings of the student model. In particular, we analyze the diversity and distribution of the pool of seed examples in an embedding space derived from the student model. Our empirical evaluation shows a strong correlation between the density of examples within a specific neighborhood and the accuracy of predictions on examples drawn from that region.  Building on this insight, we present a targeted pipeline for embedding-based sampling that samples new points in sparse regions of the embedding space to enhance data diversity. Our experimental evaluation shows that it consistently improves performance across LLMs and benchmarks. The key contributions of this paper are threefold:
\begin{enumerate}
\item An embedding-based SDG targeted to improve data diversity and quality of a specific student model.
\item An analysis of data diversity of the pool of seed examples in an embedding space derived from the student model.
\item An experimental evaluation that shows how our proposed approach consistently improves performance on different models and datasets.
\end{enumerate}
The remainder of the paper is organized as follows. After formally describing the problem  of targeted Synthetic Data Generation in section~\ref{sec:problem_statement}, we introduce our embedding based targeted SDG in section~\ref{sec:methods}. We then report, in section~\ref{sec:evaluation}, the results of our analysis of data diversity in the embedding space and the experimental evaluation of our proposed approach on two different small LLMs and on two different math datasets. After reviewing prior works in section~\ref{sec:relatedwork}, we concluded in section~\ref{sec:conclusion}.

\section{Problem Statement}
\label{sec:problem_statement}
Prior works have studied synthetic data generation with very limited to no  consideration for the target model that will eventually be fine-tuned on the generated data. In this paper, we study the problem of synthetic data generation specifically aimed at addressing the shortcomings of a given student or target model, denoted $\mathcal{SM}$. 

\begin{definition}[Generator]
    We define a targeted synthetic data generator as a function $\mathcal{F}$ that takes as input a pair ($\mathcal{D}$, $\mathcal{SM}$), where
    \begin{itemize}
        \item $\mathcal{D}=\{(x, y)\}$ is a labeled dataset made of pairs consisting of a natural language text input $x$ and a natural language output label $y$ that should be produced as a response to the problem or task described by $x$
        \item $\mathcal{SM}$ is a Large Language Model (LLM) fine-tuned on $\mathcal{D}$ from a base LLM $\mathcal{BM}$.
    \end{itemize}
    It returns a new labeled dataset $\mathcal{D'} = \mathcal{F}$($\mathcal{D}$, $\mathcal{SM}$) such that a model  $\mathcal{SM'}$ fine-tuned on $\mathcal{D'}$ from the same base model $\mathcal{BM}$ performs, on average, better than $\mathcal{SM}$ w.r.t. to some performance metrics (e.g., accuracy) on validation datasets.
\end{definition}

\section{Embedding-based SDG Method}
\label{sec:methods}

\subsection{Overview}

As indicated in the problem statement, the two key inputs to our embedding-based Synthetic Data Generation (SDG) method are a labeled training dataset $\mathcal{D}$ and a target model $\mathcal{SM}$ fine-tuned on $\mathcal{D}$. The main goal is 
 to first identify, in the embedding space, regions where the target model $\mathcal{SM}$ performs poorly. In the experimental evaluation section ~\ref{sec:evaluation},  we show that regions with a low density of examples from the training dataset $\mathcal{D}$ correlate with regions where the model $\mathcal{SM}$ performs poorly. Our method targets those sparse regions to generate synthetic examples in order to increase their density. 

\begin{figure*}
    \centering
    \includegraphics[width=0.9\linewidth]{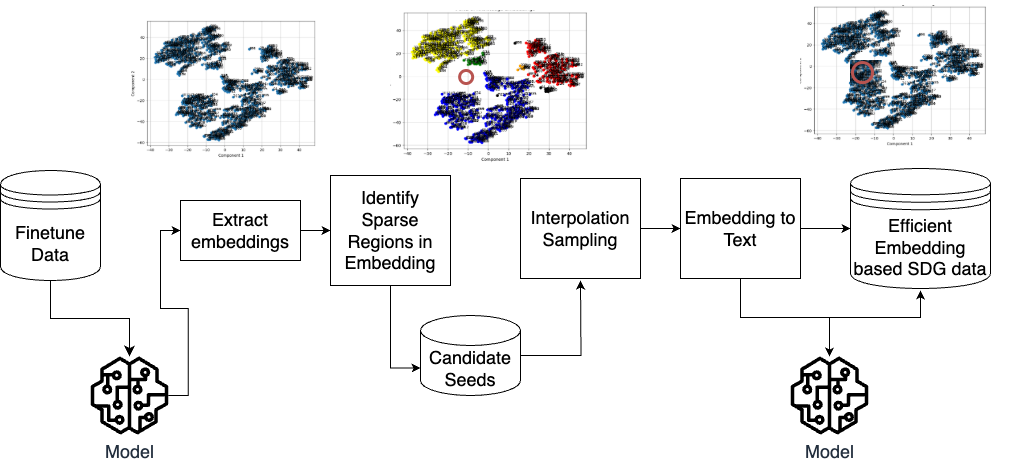}
    \caption{Pipeline to perform targeted embedding driven SDG}
    \label{fig:embedsdg_arch}
\end{figure*}

The overview of our approach is depicted in Figure~\ref{fig:embedsdg_arch}. It consists of the following key steps:
\begin{enumerate}
    \item Computation of the embedding of each example in the labeled dataset $\mathcal{D}$. $\mathcal{E}$ denotes the embedding space, and $e$ denotes the function that computes the embedding of an example in $\mathcal{D}$
    \itemsep-0.5em
    \item Identification of sparse regions of $\mathcal{E}$ (i.e., regions of $\mathcal{E}$ with low density of embeddings of examples in the labeled dataset $\mathcal{D}$). 
    \item For each identified sparse region $l$, selection of two points in $l$ that correspond to the embeddings of examples in the labeled dataset $\mathcal{D}$. We refer to those selected examples as seed examples. $s$ denotes the seed selection function. 
    \item Interpolation of the $2$ selected seed examples whose embeddings are in the sparse region $l$ to produce a new embedding vector that has a high probability of belonging to the same sparse region $l$. $i$ denotes a function that performs such interpolation and is described in more detail in  section~\ref{sec:interpolation}
    
    \item Decoding of the new embedding produced in step 4 into a natural language text using a decoding function $d$.
    \item Generation of a new synthetic data example using a Teacher LLM $\mathcal{TM}$ prompted with a prompt $\mathcal{P}_g$, the $2$ selected seed examples from step 3, and the natural language decoding of the new embedding produced at step 4.
\end{enumerate}
Let $l$ be a low density area of $\mathcal{E}$, 
$s$($l$) selects a pair of seed examples of $\mathcal{D}$ whose embeddings are in $l$. Formally, a new synthetic data point is generated by 
\begin{equation}
\mathcal{TM}( \mathcal{P}_g;  s(l);  [d(i(s(l)))]  ) 
\label{eq:sdg}
\end{equation}
where ``;'' denotes the list concatenation operator; $\mathcal{TM}$ is Teacher Model with a prompt $\mathcal{P}_g$; $d$ is the decoding function from the embedding space to natural language; $i$ is the interpolation function that combines $2$ examples in $\mathcal{D}$ and returns a new embedding in $\mathcal{E}$ (more on $i$ in section~\ref{sec:interpolation}).

Given the pair ($\mathcal{D}$, $\mathcal{SM}$), our targeted synthetic data generator function $\mathcal{F}$ is implemented by repeatedly generating new synthetic data examples (i.e., invoking the formula~(\ref{eq:sdg}) above) on sparse regions $l$ of the embedding space $\mathcal{E}$.

In the remainder of this section, we provide a detailed description of the key steps of the proposed embedding-based SDG approach.

\subsection{Embedding Computation}
\label{sec:embedding}

We first need to embed each training example in $\mathcal{D}$ in an embedding space $\mathcal{E}$. 
Given an input sequence $t$ of $m$ tokens, a transformer-based LLM~\cite{vaswani2017attention} $\mathcal{SM}$ computes, in its embedding layer, $m$ embedding vectors of dimension $N$ and an attention weight (i.e., a $m$-dimension vector where each component indicates the relative importance of each input token). Let $\mathcal{SM}^{e}(t)$ denote the $ N \times m$ matrix representing those $m$ $N$-dimension embeddings, and $\mathcal{SM}^{w}(t)$ denote the $m$-dimensional attention weight vector. We could use the weighted sum of the token embeddings as the final embedding of the input sequence $t$ computed as shown below:
\begin{equation}
    \mathcal{SM}^{e}(t) \times \mathcal{SM}^{w}(t)
    \label{eq:weightsum}
\end{equation}
 where `$\times$' is the matrix multiplication operator. However, this simple approach has two important shortcomings. First, it requires significant memory resources as $N$ is typically larger than 4000. Second, as shown in ~\cite{tyshchuk2023isotropy}, the embedding space of transformer-based models is typically not isotropic. In other words, not all dimensions are equally important. To alleviate these two issues, we perform a further step of dimensionality reduction by applying well known techniques such as PCA~\cite{jolliffe2002principal}, TruncatedSVD~\cite{hansen1987truncated}, t-SNE~\cite{van2008visualizing}, etc. The final embedding $e(t)$ of the input sequence $t$ is a $K$-dimension vector computed as 
\begin{equation}
e(t) = \text{dim\_red[K]}(\mathcal{SM}^{e}(t) \times \mathcal{SM}^{w}(t))
\end{equation}
where dim\_red[K] is a dimensionality reduction function that reduces the dimension from $N$ to $K$. In the remainder of this paper, to simplify the presentation, we consider only the cases where $K$ is 2 or 3.  



\subsection{Identifying sparsity}
\label{sec:sparsity}
\begin{figure*}
    \centering
    \includegraphics[width=1\linewidth]{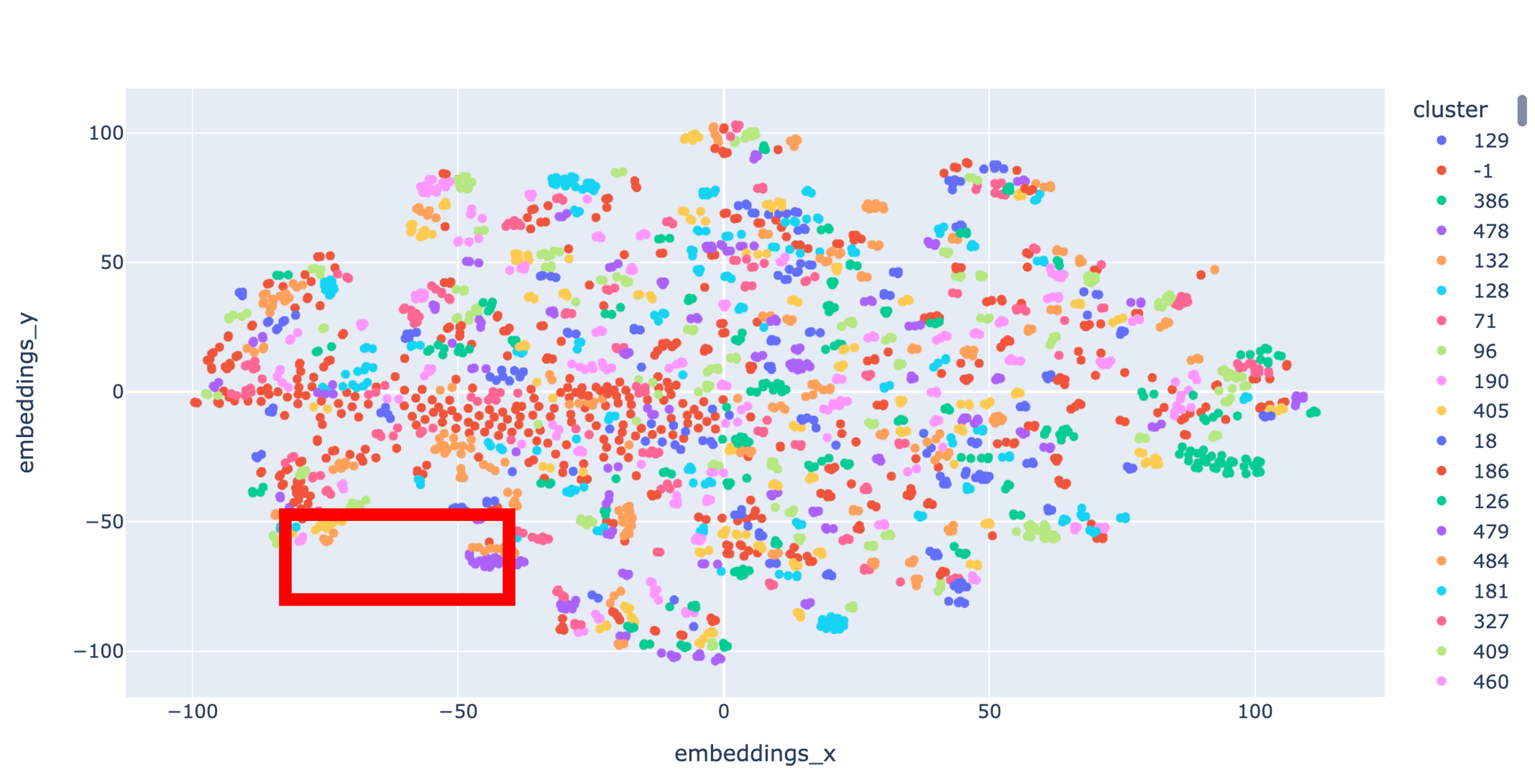}
    \caption{Red box represents potential gap (sparsity) in embedding space}
    \label{fig:gap-identification}
\end{figure*}

On visualizing the embeddings (using t-SNE~\cite{van2008visualizing}), it is seen that the data $\mathcal{D}$ that is used to fine-tune the model $\mathcal{SM}$ is not evenly distributed through the embedding space. Some areas are dense, containing samples from similar topics, while others are sparse. On clustering the data, and extracting topics for each cluster, we can identify the "topic" of the region which is dense and sparse. For each model, the embedding space distribution will be different for the same data. Each model has its own sparsity, in certain areas depending on how it embeds the data (using its embedding function $\mathcal{SM}^e$ and its attention weight function  $\mathcal{SM}^w$ as shown in expression~(\ref{eq:weightsum})) . For example, for math reasoning, Figure \ref{fig:gap-identification} shows the distribution of Meta-Math-QA data for Granite 3 ~\cite{granite31}, which was used to do supervised finetuning of Granite 3 8b code instruct.

In the $K$ dimensional space $\mathcal{E}$ of the embeddings that were produced after dimensionality reduction, the space that we consider in order to identify sparsity is the space $\mathcal{E}'$  ($\mathcal{E}' \subset \mathcal{E}$) of the embeddings which fall under the ``boundaries'' of examples in $\mathcal{D}$. i.e. for $K=2$, the space $\mathcal{E}'$  of embeddings that we consider involves the grid where the top and bottom boundaries fall at the highest and lowest value in height, and the right and left boundaries are picked as the lowest and highest values of $\{e(t) | t \in \mathcal{D}\}$ in the width dimension. Given this space $\mathcal{E}'$, in order to identify sparsity, a grid $G$ with width $w$ and height $h$ is picked, and parses through the space $\mathcal{E}'$  as a sliding window. The selection of $w$ and $h$ depends on the distribution of the samples in  $\mathcal{E}'$, and this distribution varies from model to model, on the same dataset, depending on the LLM $\mathcal{SM}$'s embedding function. Given this grid $G$, a threshold $T$ is picked depending on the density of the grids throughout $\mathcal{E}'$. Any regions which fall under this threshold $T$, i.e. any region with number of samples in grid less than $T$ is considered as a candidate sparse region $l$, from which seed examples are to be picked. In order for a region to be considered as a candidate sparse region $l$, only non-zero grids are picked, and ``empty'' regions are skipped past. For example, in the embedding space shown in Figure \ref{fig:gap-identification}, the regions in the corners are typically empty. 
These regions are identified by multiple consecutive empty grids, and are not considered as part of the sparse regions. However, when $G$ falls in an area where there are samples around the region, but there are fewer than $T$ samples, this is considered as a candidate sparse region $l$.

\begin{figure}
    \centering
    \includegraphics[width=0.9\linewidth]{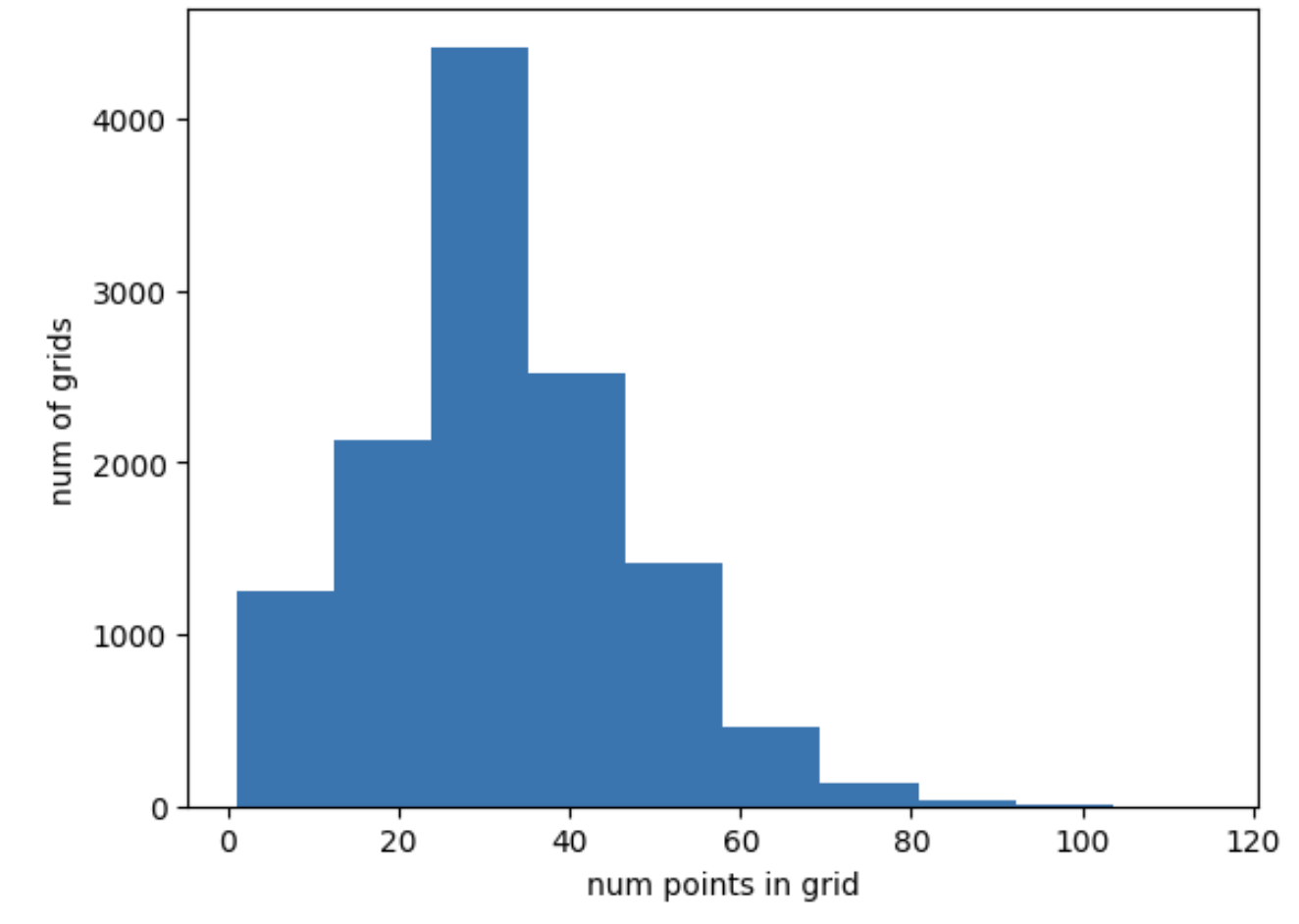}
    \caption{Grid density distribution of embedding space}
    \label{fig:threshold_selection}
\end{figure}

In order to pick $T$, we can work backwards from the density of the grids in the embedding space, for the chosen $w$ and $h$. Figure \ref{fig:threshold_selection} shows the distribution of grid density for Granite 3, where the X-axis represents the number of points in grid, and y-axis represents the number of grids which fall under this bucket. It can be seen that most grids contain 30-50 samples in the grid, with the distribution waning on the lower and higher side. Here, for a threshold $T=10$, about 1000 grids will be identified as sparse regions $l$, where the number of points in those grids fall below $T=10$.

\begin{figure}
    \centering
    \includegraphics[width=1\linewidth]{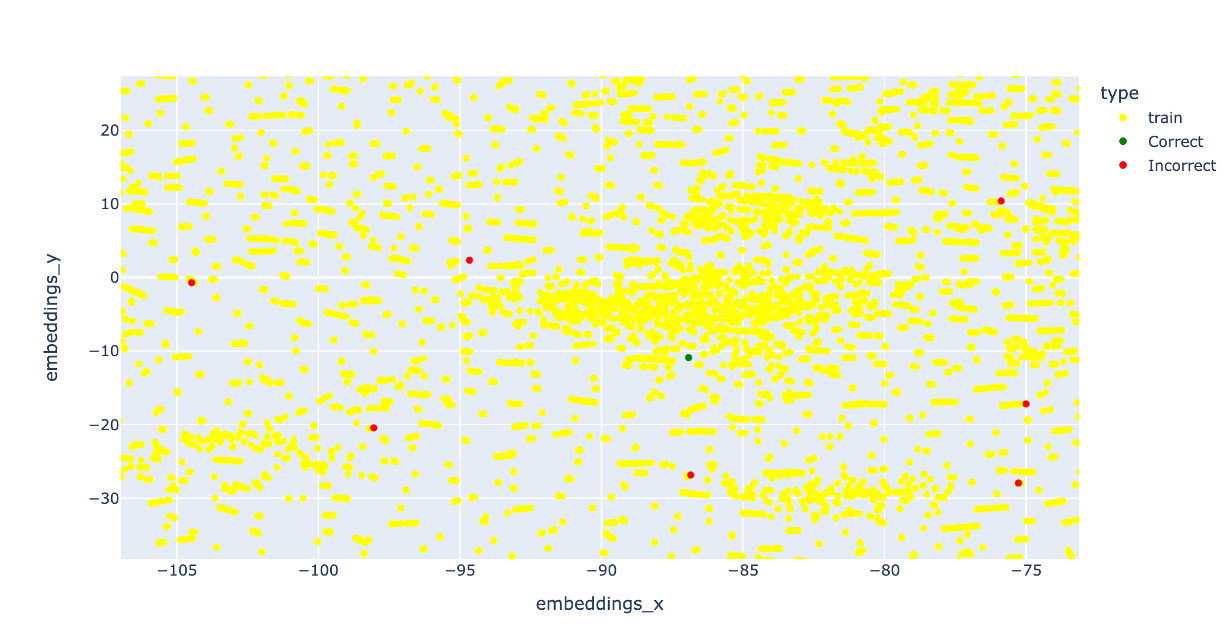}
    \caption{Motivation: Correlation between performance and density in embedding space}
    \label{fig:correlation-performance-density}
\end{figure}
\subsection{Seed Example Selection in Embedding Space}
\label{sec:seed_selection}
Once we have identified a set of sparse regions (rectangles in 2D or rectangular prisms in 3D), 2 data points from existing training data $\mathcal{D}$ are selected from the opposing sides (resp. surfaces) in 2D (resp. 3D) of each sparse region. For example, in 2D, we randomly select points from opposing sides: from either top, bottom; or right and left sides. If a data point does not exist on a side (resp. surface), we randomly select a data point close to the side (resp. surface). The hypothesis is that since there is sparsity in this region, the model is lacking in knowledge in this specific topic. So, selecting seeds from this sparse region and generating synthetic data would translate to increasing the density of this region in the embedding space, making the model more confident in this region.

\subsection{Interpolation of Selected Seed Examples}
\label{sec:interpolation}
Let $t_1 \in \mathcal{D}$ and $t_2 \in \mathcal{D}$ be two selected seed examples that are sequences of $m_1$ and $m_2$ tokens respectively. Let $m = \text{max}(m_1, m_2)$ .  Given $t_1$ and $t_2$, the interpolation function $i$ first averages their weighted embedding sequences: $\mathcal{SM}^e(t_k) \cdot \mathcal{SM}^w(t_k) $, where $\mathcal{SM}^e(t_k)$ is the $N \times m_k$ matrix of the embeddings of the $m_k$ tokens of $t_k$ computed by the model embedding layer and $\mathcal{SM}^w(t_k)$  is the $m_k$-dimension vector of the attention weights as explained in section~\ref{sec:embedding} and $\cdot$ is the element-wise matrix multiplication (with broadcasting of elements of the $m_k$-dimension vector $\mathcal{SM}^w(t_k)$ to make it a $N \times m_k$ matrix).   This results in the $N \times m$-dimension matrix $\text{avg}(\{t_1, t_2\})$ computed as follows:  
\begin{equation}
\text{avg}(\{t_1, t_2\}) = \frac{\sum_{k=1}^{2}\text{pad}_m(\mathcal{SM}^e(t_k) \cdot \mathcal{SM}^w(t_k))}{2}
\label{eq:avg}
\end{equation}
where $\text{pad}_m$ performs up to $m$ zero-padding along the columns of its input matrix. Since $\text{avg}(\{t_1, t_2\})$ has $m$ columns of $N$ elements, where $N$ is  the original embedding  dimension of the LLM $\mathcal{SM}$,  it can be passed through the rest of the LLM's pipeline to regenerate the natural language from the embeddings (see the next section~\ref{sec:decoding} for more detail).

The final step is to embed $\text{avg}(\{t_1, t_2\})$ in the $K$-dimension embedding space $\mathcal{E}$ using formula (3), where all the columns of the matrix have the same attention weight $1$ (as the original attention weights of $t_1$ and $t_2$ are already factored in the computation of $\text{avg}(\{t_1, t_2\})$ as shown in equation~(\ref{eq:avg})):

\begin{equation}
    i(\{t_1, t_2\}) = \text{dim\_red[K]}(\text{avg}(\{t_1, t_2\}) \times [1]^m)
    \label{eq:interpolation}
\end{equation}
If the dimensionality reduction function $\text{dim\_red[K]}$ is linear (e.g., PCA), then $i(\{t_1, t_2\})$ must be the mid-point of the segment [$e(t_1)$, $e(t_2)$] (as $\text{avg}(\{t_1, t_2\}) \times [1]^m$ is the mid-point of the segment [$\mathcal{SM}^{e}(t_1) \times \mathcal{SM}^{w}(t_1)$, $\mathcal{SM}^{e}(t_2) \times \mathcal{SM}^{w}(t_2)$] before dimensionality reduction), which ensures that $i(\{t_1, t_2\})$ is always in the same rectangular (or rectangular prism) region as $e(t_1)$ and $e(t_2)$. For non-linear dimensionality reduction function such a guaranty does not exist in general. However, for $t-SNE$ (used in our experiments), by construction, there is a higher likelihood that $i(\{t_1, t_2\})$ ends up closer to $e(t_1)$ and $e(t_2)$, and thus in the  same rectangular (or rectangular prism) region as them. 

\begin{figure}
    \centering
    \includegraphics[width=1\linewidth]{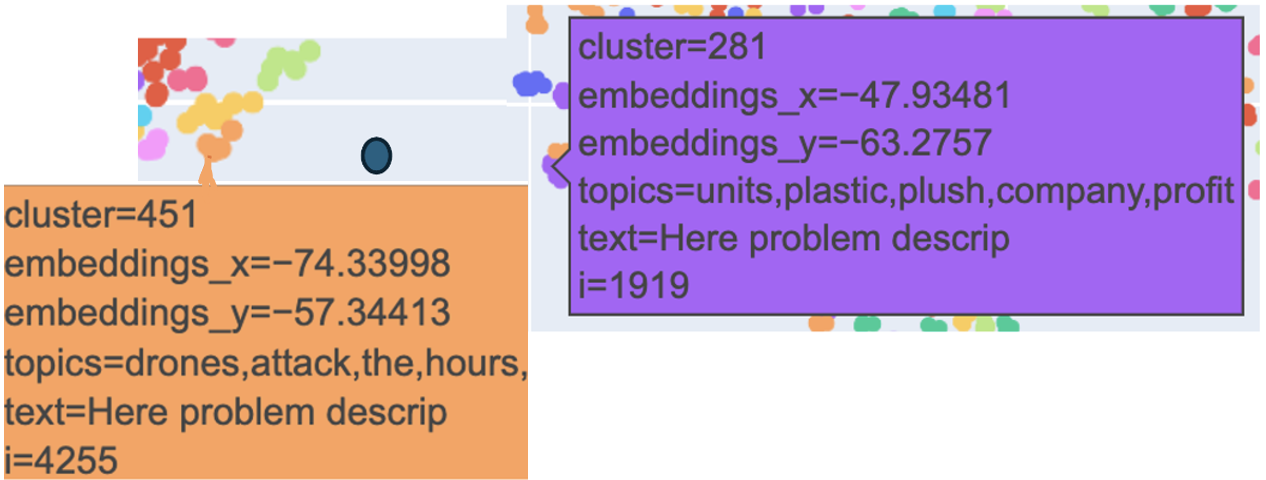}
    \caption{Interpolation: select area in sparsity between two topics}
    \label{fig:interpolation_intermediate}
\end{figure}

Figure \ref{fig:interpolation_intermediate} shows a sample sparse region where interpolation between two edge points in sparsity leads to a synthetic data example lying close to the middle of the two in the embedding space.

\subsection{Decoding of Interpolated Examples}
\label{sec:decoding}

We now describe the decoding function $d$ that regenerates text from the embeddings created by the interpolation function $i$. For an input sequence $t$ of $m$ tokens, the transformation performed by the LLM $\mathcal{SM}$ can be decomposed in two steps: the embedding step, denoted $\text{Emb}$, that computes the embedding sequence $\mathcal{SM}^e(t)$ and the attention weights $\mathcal{SM}^w(t)$ as explained in section~\ref{sec:embedding} and the generation step, denoted $\text{Gen}$, that generates the final text from the input embedding: $\mathcal{SM}(t) = \text{Gen}(\text{Emb}(t))$

The decoding of the interpolation $i(\{t_1, t_2\})$ of two examples $t_1$ and $t_2$ is done by prompting the LLM $\mathcal{SM}$ with the  decoding prompt $\mathcal{P}_d$ . Figure~\ref{fig:copy_prompt} shows a simplified version of the prompt $\mathcal{P}_d$ (the detailed and complete prompt is in listing ~\ref{prompt_p_d} in Appendix~\ref{sec:prompts}).

\begin{figure}[hbtp]
\small
    \caption{Simplified Decoding Prompt $\mathcal{P}_d$}
    \label{fig:copy_prompt}
    \begin{lstlisting}[frame=single] 
Please copy or rephrase the input text.
Use the following format:
    Input: the input text to copy
    Output: the copy of the input text
Your response should only include 
the answer. 
Do not provide any further explanation.
Here are some examples:
{Examples}
Input:
    \end{lstlisting}
\end{figure}

The prompt $\mathcal{P}_d$ instructs the LLM to simply copy the provided input. However, the input is provided in the LLM's embedding space, and the decoding is done as follows:
\begin{equation}
d(i(\{t_1, t_2\})) = \text{Gen}(\text{Emb}(\mathcal{P}_d);\text{avg}(\{t_1, t_2\}) )
\label{eq:decoding}
\end{equation}
where ";" performs the concatenation of two tensors along their last dimension, and $\text{avg}$ is as defined in section~\ref{sec:interpolation}.



\subsection{Final Generation based on Interpolated and Selected Seed Examples}
Given two selected seed examples $t_1$ and $t_2$ from $\mathcal{D}$ and  the text decoding $d(i(\{t_1, t_2\}))$ of the interpolation of $t_1$ and $t_2$, a teacher LLM $\mathcal{TM}$ is prompted with a prompt $\mathcal{P}_g$ to generate a new text example with a question and answer in the required format: \\
\begin{equation}
\mathcal{TM}( \mathcal{P}_g; [t_1, t_2];  [d(i(\{t_1, t_2\})]  ) 
\label{eq:stg2}
\end{equation}
where ``;'' denotes the list concatenation operator. The prompt $\mathcal{P}_g$ is provided in listing~\ref{prompt_p_g} in Appendix~\ref{sec:prompts}.

\section{Experimental Evaluation}
\label{sec:evaluation}



\begin{table*}[!ht]
    \caption{Experimental results comparing accuracy for random vs Embedding-driven seed selection on math reasoning}
    \label{experiments}
    \centering
    \begin{tabular}{|l|c|c|c|c|c|c|c|}
    \hline
        \rowcolor[HTML]{E5E5E5} 
        \multirow{2}{*}{\bfseries Model} & \multicolumn{3}{c|}{\bfseries Random Seed Selection} & \multicolumn{3}{c|}{\bfseries EmbedSDG} & \multirow{2}{*}{\bfseries Base Model} \\ \cline{2-7}
        & \bfseries 500 & \bfseries 1000 & \bfseries 4500 & \bfseries 500 & \bfseries 1000 & \bfseries 4500 & \\ \hline
        
        \multicolumn{8}{|c|}{\bfseries Math - GSM8k} \\ \hline
        Granite 3 8b code instruct & 0.761 & 0.761 & 0.773 & 0.782 & 0.79 & \bfseries 0.79  & 0.55 \\ \hline
        Granite 3.1 8b instruct & 0.786 & 0.786 & 0.785 & \bfseries 0.824 & 0.806 & 0.81 & 0.74 \\ \hline
        Mistral 7B & 0.354 & 0.558 & 0.723 & 0.62 & 0.725 & \bfseries 0.746 & 0.354 \\ \hline

        \multicolumn{8}{|c|}{\bfseries Math - MATH} \\ \hline
        Granite 3 8b code instruct & 0.225 & 0.23 & 0.24 & 0.249 & 0.266 & \bfseries 0.2822 & 0.18 \\ \hline
        Granite 3.1 8b instruct & 0.28 & 0.357 & 0.357 & 0.342 & 0.321 & \bfseries 0.3612 & 0.2 \\ \hline
        Mistral 7B & 0.214 & 0.225 & 0.229 & 0.244 & 0.244 & \bfseries 0.248 & 0.11 \\ \hline
    \end{tabular}
\end{table*}


\subsection{Datasets and Benchmarks}
In order to establish the usefulness of our method, we focus on one task of Math Reasoning in LLMs. 
We pick 3 comparable models in the same parameter range, Granite 3 8B Code Instruct ~\cite{granite2024granite}, Granite 3.1 8b Instruct 
\cite{granite31} and Mistral 7B \cite{jiang2023mistral7b} and consider them as the target models in our experiments. We consider MetaMathQA \cite{metamath} as the base dataset to select seeds from, as it has been used in supervised finetuning for the 3 models selected base models. MetaMathQA which is a publicly available dataset for mathematical problem solving, containing about 400K examples (as input and output pairs). The experiments on table \ref{experiments} show the performance of the 3 models on the test split of 2 popular benchmark datasets for mathematical reasoning: GSM8K \cite{gsm8k} and MATH \cite{math}. We pick these models since a prerequisite for our method is that the pool of seed examples need to have been used in finetuning the model, and not many other models disclose this information. The results show that our method consistently outperforms the random seed selection in every model for every benchmark dataset. 

\subsection{Experimental setup}
Table \ref{experiments} shows the comparison of the accuracy of the 3 target models : Granite 3 8b code instruct, Granite 3.1 instruct and Mistral 7B on GSM8K and MATH, each of which have been finetuned using the method and sample size denoted in the columns. The column headers denote the final number of examples used to finetune the respective SMs, according to the method of "Random seed selection" or our pipeline of ``Embedding driven SDG'', denoted as ``EmbedSDG''. 

For the "random seed selection" case, we pick the seed examples randomly irrespective of their location in the embedding distribution, and use the seed examples to generate synthetic data based on the seed examples (prompt in Appendix \ref{prompt_p_b}). Here, it is a 1:1 mapping of number of seed examples, and number of synthetic data examples generated. For instance, the first column represents 500 seed examples randomly selected from the finetuning data D(MetaMathQA), which gave rise to 500 synthetic data samples generated by prompting a teacher large language model (Mistral-Large \cite{mistral_large}), and similarly 1000 and 4500 examples generated by randomly sampling from the finetune data and generating examples. 

The second subdivision of "EmbedSDG" which has 3 columns underneath: 500, 1000, 4500 refers to our method of producing synthetic data, and the number of examples are the final number of examples generated at the end of our pipeline. Here, for each synthetic data sample that is generated, as explained in the Methodology section \ref{sec:methods}, two seed samples are picked from a sparse region, interpolated to generate a new sample in the targeted sparse region, decoded, and the decoded text is further solidified into a legitimate example by prompting a teacher large language model (Mistral-Large) with the prompt specified in the Appendix \ref{prompt_p_d}. So, the column represents the number of examples that were generated at the end of the pipeline considering the appropriate number of seed pairs and interpolated data. For instance, the 500 column represents the accuracy of the target model in the respective rows finetuned with 500 synthetic data samples generated via our pipeline (embedSDG), and so on. The final column represents the accuracy results of the base model without any further finetuning.  

Finetuning on each of the target models was done using a single node of 1 $A100\_80GB$ GPU with 80GB of RAM for a total of 5 epochs using Low Rank Adaptation (LoRA) \cite{hu2021lora}.

\subsection{Evaluation results (Efficiency of embedding based SDG)}
Table \ref{experiments} shows the accuracy results of the target models finetuned on both random seed selection driven SDG and Embedding based seed selection driven SDG, with the best accuracy for each target model on each dataset represented in \textbf{bold}. The results show that our method EmbedSDG consistently performs the best across all models and across all benchmarks, over RandomSDG and the base models. In all cases, the method improves performance significantly as compared to the base model, with upto \textbf{39\%} improvement for Mistral7B on GSM8K, and upto \textbf{16\%} improvement from base model for Granite 3.1 instruct for the MATH dataset. \\
The improvement in performance over RandomSDG is especially noticeable in the case where the number of examples is low: for instance, Mistral7B on GSM8K observes almost a 2X improvement (\textbf{0.62} on EmbedSDG as opposed to \textbf{0.35} on RandomSDG) over the RandomSDG pipeline, with just 500 examples. As expected, the improvement over RandomSDG becomes lesser pronounced with an increase in the number of seed examples: this is because, as we increase the number of seed examples that we are sampling, we foray into regions which becomes less sparse than the previous sample set. For instance, for EmbedSDG, the first column would be generated from 500 seed samples from the least dense (or most sparse) regions in the embedding space. As we increase this number, the sparsity is less pronounced.\\
However, the EmbedSDG method still consistently leads to the most useful synthetic data since the highest performance in terms of accuracy are for the \textbf{EmbedSDG 4500} pipeline for all target models for both benchmark datasets GSM8K and MATH, except for Granite 3.1 for which the best performer is the target model finetuned with \textbf{500} samples generated via EmbedSDG.


\subsection{Correlation between Density  and Model Accuracy}
Our experiments demonstrate a strong correlation between the number of fine-tuning examples available for a specific region and the accuracy of the target model in that region within the embedding space. As the number of examples increases in a region, the model's performance improves, highlighting that data sparsity is a significant factor contributing to accuracy disparities across regions in the embedding space. This trend is consistently observed across all models for the benchmark datasets.

\begin{figure}
    \centering
    \includegraphics[width=0.8\linewidth]{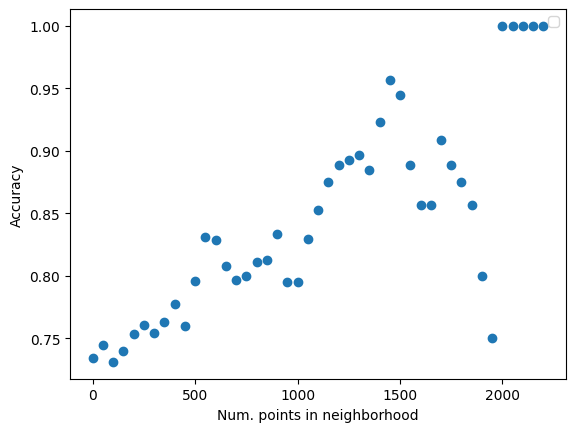}
    \caption{Correlation between density in ES and Model Accuracy}
    \label{fig:Correlation Graph}
\end{figure}

Figure \ref{fig:Correlation Graph} illustrates this linear relationship: the X-axis represents the number of points within a region in the embedding space, where the dimensionality has been reduced to 3 dimensions using PCA. The Y-axis depicts the average accuracy of the target model in that region. The positive relationship is evident from the graph and is further confirmed by statistical correlation analysis.

\subsubsection{Statistical correlation analysis}
We computed both Pearson's and Spearman's correlation coefficients to validate the strength of the relationship. Pearson's correlation coefficient was found to be \textbf{0.813}, with a p-value of \textbf{$1.0946e^{-11}$}, indicating a strong positive linear relationship and confirming statistical significance. Spearman's correlation coefficient was \textbf{0.806}, with a p-value of \textbf{$2.2910e^{-11}$}, confirming a strong monotonic relationship with an equally significant p-value. These results highlight the importance of addressing data sparsity to reduce accuracy disparities.

\section{Related Work}
\label{sec:relatedwork}
Instruction fine-tuning has improved the ability of large language models (LLMs) to generate accurate and contextually relevant outputs \cite{touvron2023llama2openfoundation, geminiteam2024geminifamilyhighlycapable, granite2024granite}. However, obtaining large, high-quality instruction datasets remains a challenge, leading to the exploration of synthetic data generation (SDG) methods, such as Self-Instruct \cite{selfinstruct} and $Evol\text{-}Instruct$ \cite{xu2023wizardlmempoweringlargelanguage}. In past, SDG has used for a target specific domains like mathematics \cite{toshniwal2024openmathinstruct1}, web development \cite{puranik-etal-2023-protege}, education \cite{bulathwela2023scalable, 2022.EDM-posters.85}, and machine learning exams \cite{10.1145/3580305.3599827}. 

While larger datasets, such as Self-Align’s 300K+ instructions and Self-Instruct’s 50K+, generally improve model performance, they also introduce issues like neural text degeneration \cite{Holtzman2020The} and data contamination \cite{li2023open}. Recent studies advocate for prioritizing content quality over quantity in fine-tuning \cite{guo2023curious, liu2023makes}. Thus, an effective instruction generation system must control the quality of the generated data. Additionally, leveraging the geometric structure of language embeddings for synthetic data generation \cite{watchful2024} has shown promise, particularly in domains like clinical text, where expert annotations are costly. For example, embedding-based techniques \cite{ebmeddingwork1} use diversity sampling from real clinical notes to guide language models, producing synthetic text that mirrors clinical syntax and vocabulary. Instruction generation is often tailored to the characteristics of the target language model, but there is limited research on generating synthetic instruction data for fine-tuning smaller models, which aligns more closely with end-user needs. 


\section{Conclusion}
By carefully navigating the embedding space of a target model for a given task, this paper introduces a novel synthetic data generation (SDG) pipeline that consistently improves
performance on different models and datasets. We empirically demonstrate that a model's performance is closely correlated with the density within the embedding search space. This observation provides valuable insights for future research, offering guidance for optimizing the synthetic data generation process to improve model performance. As part of future work, we plan to develop a multi-task embedding space to generate more complex instructions.

\label{sec:conclusion}

\section*{Limitations}
We identify two primary limitations of our work. The first is experimental in nature: our approach has been evaluated only on 3 models, on 2 datasets. All models we used have been trained and fine-tuned using data from this domain, which may limit the generalizability of our findings to other domains. This is because our approach focuses on improving the performance of a finetuned model, and most models do not disclose which datasets were used to train on. This limits the datasets and models we could experiment on, due to lack of information, and the need for using a dataset that was explicitly used in the finetuning of an instruction tuned model.

The second limitation concerns computational resources. While larger models such as 70B parameter variants demonstrate strong performance, building and deploying solutions with such models requires substantial hardware resources, making them less accessible for real-world applications in resource-constrained environments. However, the scope of our approach is mainly to improve the performance of smaller models to be more effective than large models requiring a lot more cycles of compute and resources.



\nocite{langley00}
\bibliography{custom}

\newpage
\appendix
\onecolumn
\section{Appendix}

\subsection{Prompts}
\label{sec:prompts}

 \begin{lstlisting}[frame=single,caption={Prompt to decode interpolated example $\mathcal{P}_d$},label={prompt_p_d}] 
You are a cautious assistant. You carefully follow instructions. 
You are helpful and harmless and you follow ethical guidelines and 
promote positive behavior.

Please copy or rephrase the input text.

Use the following format:
  Input: the input text to copy
  Output: the copy of the input text

Your response should only include the answer. Do not provide any 
further explanation.

Here are some examples, complete only the last one:

Input: {Example 1}
Output: {Example 1}

Input: {Example 2}
Output: {Example 2}

Input: {Example 3}
Output: {Example 3}

Input: {Example 4}
Output:{Example 4}

Now copy or rephrase the following input text. 
Do not try to solve the problem described in the input text. 
Just copy or rephrase the following input text.
Input: 
 \end{lstlisting}

    \begin{lstlisting}[frame=single,caption={Prompt to generate a new synthetic example from seeds $\mathcal{P}_g$},label={prompt_p_g}] 

You are a data generator for synthetic math reasoning problems.
You will be given two examples of math reasoning problems in a 
specific format, along with a partial synthetic example, and 
your task is to generate a new problem that creatively combines
elements of the two examples, and solidifies the partial example
into a legitimate math reasoning problem with a legitimate 
solution.


Format:

- Start the question with "### Question".
- Start the answer with "### Answer".
- Provide the final answer prefixed with "### <final_answer>".

Instructions:

- The new problem should conceptually lie in the middle of the two 
   given examples, and follow the outline of the partial example.
- Combine themes or elements from both problems to create a coherent 
   and challenging math reasoning problem. It has to be related to 
   the seed examples, and cannot be an unrelated random problem.
- Ensure that the problem adheres to the format provided.
- Do not use the same names or numbers, only use the concepts, 
   topics and problem types.

  
Seed Examples to base the new sample on:

Example 1:
### Question

{q1}

### Answer

{a1}

Example 2:
### Question

{q2}

### Answer

{a2}

Partial Example:
{interpolated_decoded_text}

Your Task:

Generate a new math reasoning problem that combines the elements of 
Example 1 and Example 2 above, and solidifies the Partial Example 
into a real math reasoning problem. The new example HAS to have 
elements from the above two seed examples, it cannot be an unrelated 
random math problem. Follow the format exactly and ensure the problem
is clear and solvable. Only respond with one new example, and preface 
the question with ### Question and the answer with ### Answer.
  

Generated example:
    \end{lstlisting}

 \begin{lstlisting}[frame=single,caption={Prompt to decode interpolated example $\mathcal{P}_d$},label={prompt_p_d}] 
You are a cautious assistant. You carefully follow instructions. 
You are helpful and harmless and you follow ethical guidelines and 
promote positive behavior.

Please copy or rephrase the input text.

Use the following format:
  Input: the input text to copy
  Output: the copy of the input text

Your response should only include the answer. Do not provide any 
further explanation.

Here are some examples, complete only the last one:

Input: {Example 1}
Output: {Example 1}

Input: {Example 2}
Output: {Example 2}

Input: {Example 3}
Output: {Example 3}

Input: {Example 4}
Output:{Example 4}

Now copy or rephrase the following input text. 
Do not try to solve the problem described in the input text. 
Just copy or rephrase the following input text.
Input: 
 \end{lstlisting}

    \begin{lstlisting}[frame=single,caption={Prompt to generate a new synthetic example from baseline seed $\mathcal{P}_b$},label={prompt_p_b}] 
You are a data generator for synthetic math reasoning problems. 
You will be given an example math reasoning problem in a specific 
format, and your task is to generate a new problem that that is 
similar to the example problem, and convert it into a 
legitimate math reasoning problem with a legitimate solution.

Format:

- Start the question with "### Question".
- Start the answer with "### Answer".
- Provide the final answer prefixed with "### <final_answer>".

Instructions: 

- The new problem should conceptually be similar to the example problem.
- Create a coherent and challenging math problem.
- Ensure that the problem adheres to the format provided.
- Do not use the same names or numbers, only use the concepts, topics and problem types.


Seed Example to base the new sample on:


Example:
### Question
{q1}

### Answer
{a1}

Your Task:

Generate a new math reasoning problem similar to the example problem 
above. The new example HAS to be similar to the example, it cannot be 
an unrelated random math problem. Follow the format exactly and ensure 
the problem is clear and solvable. Only respond with one new example, 
and preface the question with ### Question and the answer with 
### Answer.

Generated example:
    \end{lstlisting}


\subsection{LLM Licenses}
\label{sec:licenses}
Experiments described in section~\ref{sec:evaluation} were performed in a manner consistent with the license and intended use of our three target models:  Granite 3 8B Code Instruct ~\cite{granite2024granite}, Granite 3.1 8b Instruct 
\cite{granite31} and Mistral 7B \cite{jiang2023mistral7b}.

\end{document}